\documentclass[runningheads]{llncs}
\usepackage[T1]{fontenc}
\usepackage{graphicx}
\usepackage{amsmath}
\usepackage{amssymb}
\usepackage[T1]{fontenc}
\usepackage[utf8]{inputenc}
\usepackage{booktabs}
\usepackage{multirow}
\usepackage{algorithm}
\usepackage{algpseudocode}

\usepackage{booktabs}
\usepackage{multirow}

\usepackage{xcolor}

\makeatletter
\renewcommand\subsubsection{\@startsection{subsubsection}{3}{\z@}%
  {-0.8ex plus -0.2ex minus -0.2ex}
  {-0.5em plus -0.1em minus -0.1em}
  {\normalfont\normalsize\bfseries}}
\makeatother

\begin{document}
\title{Outage Detection in Self-Healing Smart Grids Using Reinforcement Learning with Spectral Graph Neural Networks}

%
%






\author{
Lihui Liu\inst{1} \and 
Mucun Sun\inst{2} \and 
Caisheng Wang\inst{1}
}

\authorrunning{L. Liu and C. Wang}

\institute{
Wayne State University, Detroit, MI, USA
\email{\{hw6926, cwang\}@wayne.edu}
\and
\email{mxs165231@utdallas.edu}
}

%
\maketitle              

\begin{abstract}
Self-healing smart grids can quickly adjust their network configuration during outages to minimize power disruptions. During an outage, several actions can be taken, such as network reconfiguration through switching operations and emergency load shedding. However, traditional machine learning methods for outage mitigation are not well suited for smart grids due to their slow response time and high computational cost.
To address these challenges, recent studies have explored reinforcement learning to automatically perform network reconfiguration. In these approaches, the control policy is typically modeled using a graph neural network (GNN). However, conventional GNNs operate in the spatial domain and may fail to capture important relationships in the frequency domain. Frequency-domain information is particularly useful for modeling global structural patterns and system-wide interactions in power networks.
In this paper, we propose a spectral graph reinforcement learning framework for outage management in distribution networks to enhance system resilience. Our model learns the optimal power restoration policy using a spectral graph neural network. We evaluate the proposed method on three modified IEEE test systems: the 13-bus, 34-bus, and 123-bus networks. Experimental results show that our approach achieves near-optimal performance in real time and generalizes well across a wide range of outage scenarios.
\end{abstract}
\section{Introduction}

For a long time, power distribution networks have been viewed simply as a link between transmission systems and end users, and their resilience has often been overlooked. However, recent analysis~\cite{jacob2024real} shows that 90\% of customer outages during extreme events are caused by failures within the distribution network itself. In addition, the increasing integration of distributed energy resources (DERs) has significantly changed the structure of the power grid. This decentralization enables local generation and creates autonomous subsystems that can operate independently of the main grid. As a result, distribution networks are no longer passive connections but active and potentially self-sustaining components of the power system.

Therefore, it is essential to develop intelligent and automated technologies for distribution networks. Today, many monitoring devices are deployed in these networks, including line monitors, fault indicators, remote-controlled switches, and reclosers. One of the most important features of a smart grid is its self-healing capability. This means the system can automatically detect faults, isolate affected areas, and restore power with minimal human intervention. By implementing intelligent control and automation, the grid can reduce power interruptions and recover more quickly after disturbances. Therefore, the goal is to design a self-healing algorithm that is autonomous, responds quickly, and can adapt to changing conditions in real time.

The smart grid typically operates as an independent entity managed by an Independent System Operator (ISO). However, coordination across multiple grids can be challenging because different ISOs may follow distinct operational rules, communication protocols, and control strategies. These differences make inter-grid operations complex, especially during large-scale disturbances that require coordinated responses.
In the face of power disruptions caused by extreme weather events or cyber-physical attacks, a self-healing distribution network (DN) must be capable of automatically detecting faulty components, isolating them, and restoring service through intelligent control algorithms. The restoration and recovery of DN operation can be achieved through various control actions, including network reconfiguration, load management, energy storage control, and reactive power resource regulation.

Among these actions, network reconfiguration is typically the first and most immediate response, as it can quickly isolate faults and restore service to unaffected areas by adjusting switch statuses. If reconfiguration alone is insufficient, load shedding may be applied to maintain system stability and prevent cascading failures. Controlling network switches is one of the most commonly used strategies in distribution systems, as it supports multiple operational objectives such as loss minimization, reliability enhancement, load balancing, and service restoration.
A comprehensive restoration strategy that is robust to diverse outage scenarios must efficiently coordinate both grid-forming and grid-feeding distributed energy resources (DERs). It should also consider all feasible reconfiguration options under operational constraints, including radiality requirements, voltage limits, and line capacity limits. Therefore, in this work, we focus on a coordinated restoration framework that integrates optimal network reconfiguration and adaptive load shedding, aiming to achieve fast, autonomous, and resilient recovery of distribution network operations under extreme conditions.

Power distribution networks are typically unbalanced and radial, meaning electricity flows in one direction from substations to customers. Their operation is already complex due to nonlinear power flow equations, and the increasing integration of distributed energy resources (DERs), such as solar panels and storage systems, has made management even more challenging.
Restoring a distribution network after an outage is a difficult optimization problem. It aims to maximize power supply while satisfying connectivity and operational constraints. This problem is NP-hard, nonlinear, and combinatorial. Existing solutions include heuristic methods, meta-heuristic algorithms, and mixed-integer programming techniques. With the growing use of DERs, researchers have also studied islanding strategies, where parts of the network operate independently from the main grid. Load management has similarly been used as an emergency control strategy. However, few studies jointly consider both grid-connected and islanded reconfiguration for outage management. The problem becomes even more complex as the number of controllable devices—such as switches, loads, and DER operating modes—increases. Although more remotely controlled devices enable greater automation, they also significantly increase computational complexity.

To address these limitations, we propose a reinforcement learning–based model that provides adaptive, real-time decision support for outage management. Reinforcement learning (RL) has recently been widely used in power system applications that require automatic and intelligent control. RL is especially suitable for solving complex, high-dimensional, and combinatorial optimization problems. It can also make decisions quickly, which is very important during outages when fast response is critical. Traditional optimization methods are often too slow for real-time decision-making.
Deep reinforcement learning (DRL) has been increasingly applied to voltage control in active distribution networks. For example, some studies use DRL to control DER inverters and static VAR compensators to maintain stable voltage levels. Other works treat distributed energy storage devices as independent agents and apply multi-agent DRL for voltage regulation, allowing the system to adapt to topology changes. Multi-agent DRL has also been used to optimally schedule DERs, storage systems, and flexible loads. In these cases, each inverter or storage device can act as an individual agent, which matches their distributed control structure.

In our model, we treat the power distribution network as a graph. The nodes represent substations, loads, or distributed energy resources (DERs), and the edges represent the power lines or transformers connecting them. Each node and edge has measurements, like voltage, current, or power demand, which are connected in complex ways.
During outages, we often need to reconfigure the network by turning lines on or off. This is difficult because many combinations are possible, and changing one part affects the rest of the network. To handle this, we use Graph Reinforcement Learning (GRL), which can control both the network connections and loads in real time.
Instead of a standard GNN, we use a Spectral Graph Neural Network. This type of network is better because it can capture both the local connections and the global structure of the network, including frequency-domain relationships. This gives a richer and more accurate representation of the network, which helps the model make better decisions.
We tested our spectral GRL model on several power networks, and it was able to control the system in real time with near-optimal performance. This makes it very effective for building self-healing power networks that can recover quickly from failures.

\section{Related Works}
\subsubsection{Optimization-based methods}
Early distribution network restoration approaches formulate the problem as mixed-integer programs, including MILP \cite{chen2018toward,pareja2022mixed,zhu2024mixed} and MISOCP \cite{xiao2018robust,wang2018using,yao2017resilience}, which guarantee optimality but are computationally expensive and poorly suited for real-time deployment. Heuristic and meta-heuristic methods such as BPSO \cite{oliveira2015optimal,lu2024multi,liu2024application} reduce solve times but may converge to suboptimal solutions and lack adaptability to dynamic network conditions.

\subsubsection{GNN-based methods} Graph learning has been studied for a long time \cite{liu2019g,liu2021neural,liu2021kompare,liu2022joint,liu2022comparative,liu2022knowledge,liu2023knowledge,liu2023knowledge2,liu2024logic,liu2024can,liu2024new,liu2024conversational,liu2025neural,liu2025few,liu2025monte,liu2025hyperkgr,liu2025mixrag,liu2024knowledge,liu2026neural,liuneural,liu2026ambiguous,liu2025unifying,liu2026dynamic,liu2026accurate,liu2026symbolic,liu2026morgan,liu2026prompt,liu2026neural2,wu2026mixture,liu2026multi}. 
Graph neural networks have been applied to power system control by capturing relational structure through neighborhood aggregation. CompGCN-style encoders \cite{vashishth2019composition,shan2025fault,dai2025fault,yang2025efficient} and their variants incorporate edge-type information for topology-aware representations. However, standard GNNs rely on local message passing and fail to capture global network structure, which is critical for distribution networks where switching one line affects voltage profiles across the entire feeder. Spectral GNN approaches address this by operating in the frequency domain of the graph Laplacian, encoding both local and long-range dependencies simultaneously \cite{defferrard2016convolutional,kipf2017semi,xiao2023automatic,zhou2025meteorological}.

\subsubsection{Reinforcement learning-based methods}
RL has been increasingly applied to distribution network reconfiguration and voltage control. DQN and PPO-based agents have been used to learn switching policies \cite{li2022learning,zhao2025review} and DER dispatch strategies \cite{thattai2024hierarchical,gao2023energy,ikram2025networked}. Multi-agent RL extends this to distributed control of inverters and storage devices. However, most RL methods use MLP-based encoders that ignore graph topology, leading to constraint violations and poor generalization across network sizes.

\subsubsection{Graph reinforcement learning for power systems}
Combining GNNs with RL enables topology-aware decision-making. GCAPS-GNN \cite{jacob2024real,xinyi2019capsule} applies capsule-based graph convolutions within a PPO framework for outage management, demonstrating improved feasibility over MLP baselines. Nevertheless, GCAPS relies on spatial aggregation and does not exploit the spectral structure of the network graph. Our approach instead employs spectral graph filtering within the RL policy, capturing global frequency-domain topology to support scalable, real-time restoration across networks of varying size.

\section{Methodology}

\subsection{DN Representation as a Graph}
Outage management in distribution networks (DNs) using switching control can be largely viewed as a task of learning the associated network topology, which motivates the reformulation of the problem in graph-theoretic terms. Consequently, we represent the DN as a graph $G=(N,E)$, where $N$ is the set of nodes interconnected by a set of edges $E$. The nodes in the graph represent the buses in the DN, including the substation, load, DER, and zero-power injection buses. The edges represent the distribution lines and inline transformers, consisting of both switchable (sectionalizing and tie) and non-switchable lines. 

The node variables comprise both forecasted or estimated variables and measured variables. These include estimated or forecasted values for active power demand (or generation), reactive power demand (or generation), and the three-phase voltage measured at each bus. The edge variable considered is the measured power flow through the branches. To obtain these measured signals, we utilize a power flow simulator in our synthetic approach. Network reconfiguration in the graph domain essentially involves determining the status (open or closed) of the switchable edges in the DN. Emergency load shedding at the primary DN level is indicated using a binary variable associated with the nodes representing switchable loads.

\subsection{A Markov decision process over graphs}
The emergency response during outages in the DN is formulated as a Markov Decision Process (MDP) in the graph domain, denoted as $M=(S,A,P_{tr},R)$. Here, $S$ represents the state, $A$ the action, $P_{tr}$ the transition probability, and $R$ the reward. The state $S$ is composed of relevant observations from the DN that represent the current operating condition of the network. It includes node variables, edge variables, network topology, and other system variables, denoted as 
\[
S = [P_N^d, Q_N^d, P_N^g, Q_N^g, V_N, V_{\text{viol}}, l_E, T, E_{\text{supp}}, O, \mu].
\] 
In this representation, $P_N^d$ and $Q_N^d$ are the estimated or forecasted active and reactive power demand at the nodes, while $P_N^g$ and $Q_N^g$ correspond to the active and reactive power generation. The three-phase voltage measured at the buses is $V_N$, and $V_{\text{viol}}$ indicates voltage violations. The edge variable $l_E$ captures the power flow through network branches. The operating topology of the network is denoted by $T$, the total energy supplied by $E_{\text{supp}}$, and $O$ represents the outage scenario, including multi-line failures and switch outages. The inoperability of outage switches is addressed using a masking mechanism, represented by $\mu$.

The action $A$ represents control operations for emergency response, including switching lines and load shedding. The action space is defined as 
\[
A = [\delta_{sw_1}, \delta_{sw_2}, \dots, \delta_{sw_{N_S}}, \delta_{ld_1}, \dots, \delta_{ld_{N_L}}].
\] 
Here, $N_S$ is the number of switchable lines, including sectionalizing and tie lines, and $N_L$ is the number of switchable loads. Line switching is represented by a binary variable $\delta_{sw}$, where $0$ and $1$ indicate open and closed switches, respectively. The status of loads is represented by a binary variable $\delta_{ld}$, where $1$ denotes served and $0$ denotes shed.

The transition probability $P_{tr}$ captures the dynamic response of the network under emergency actions, expressed as 
\[
P(s_{t+1} \mid s_t, a_t),
\] 
representing the probability of transitioning from state $s_t$ to $s_{t+1}$ given action $a_t$. The agent learns this transition through interactions with the environment.

The reward $R$ guides the learning agent to take optimal control actions for outage mitigation and is formulated as 
\[
r(s, a) =
\begin{cases}
E_{\text{supp}} - V_{\text{viol}}, & \text{if } C_{\text{viol}}=0\\
0, & \text{otherwise}
\end{cases}
\] 
where $C_{\text{viol}}$ accounts for ill-conditioned sections of the network due to outages or switching actions. The reward reflects the objective of maximizing energy supply while minimizing voltage violations. The DN may consist of multiple independent sections, each containing active components (transformers, regulators, generators, loads, etc.) with corresponding state variables. Isolation of these components from a robust slack (substation) can render the network ill-conditioned, making it difficult to maintain nodal power balance within acceptable tolerance. Non-convergence of the power flow in such cases is identified by solver flags, and the reward is set to zero to indicate the indeterminate impact of switching.

Voltage violations for each bus $i\in N$ beyond its upper limit $V_{\max}$ and lower limit $V_{\min}$ are evaluated as
\begin{align}
\Delta V_i^{\max} &= 
\begin{cases} 
\sum_{j \in \phi} (V_{ij} - V_{\max}), & V_{ij} > V_{\max} \\ 
0, & \text{otherwise} 
\end{cases} \\
\Delta V_i^{\min} &= 
\begin{cases} 
\sum_{j \in \phi} (V_{\min} - V_{ij}), & V_{ij} < V_{\min} \\ 
0, & \text{otherwise} 
\end{cases} \\
V_{\text{viol}} &= \frac{\sum_{i \in N} (\Delta V_i^{\max} + \Delta V_i^{\min})}{3 |N|}
\end{align}
where $\phi$ denotes the set of phases for each bus. All voltages and energy quantities are expressed in per-unit (pu) relative to base voltage $kV_{\text{base}}$ and base power $MVA_{\text{base}}$.

The outage management tool is applied in DNs where the distribution system operator (DSO) or substation agent regulates power balance and resource control for safe operation. In this study, test feeders feature a single substation supplying loads while integrating DERs, motivating a centralized control approach. Multi-agent system (MAS) approaches may be unsuitable for reconfiguration because they rely on wide-area measurements, especially in networks with limited observability and local information. MAS approaches, although computationally efficient, may struggle to consistently achieve optimal results. 

In contrast, our approach uses a Spectral Graph Neural Network (Spectral GNN) as the policy model for reinforcement learning. The Spectral GNN captures both spatial and frequency-domain relationships, allowing accurate representation of network topology and combinatorial interactions between switching and continuous variables. This enables near-optimal decisions for network reconfiguration and load shedding under diverse outage scenarios. By integrating global (wide-area) and local network information, the Spectral GNN facilitates real-time, scalable, and resilient DN outage management, consistent with realistic DN control architectures. While the control of smart grids may evolve toward distributed architectures in the future, the current framework assumes an established centralized DSO agent, providing a practical and implementable solution for real-world DN outage management.

\section{Learning Architecture}

The learning architecture employs a policy gradient-based Graph RL algorithm where the policy network is derived from a Spectral Graph Neural Network (Spectral GNN). Each node $i$ in the DN graph has properties such as active/reactive power demand and generation, and three-phase voltage measurements, denoted as 
\[
\gamma_i = [P_i^d, Q_i^d, P_i^g, Q_i^g, V_i].
\] 
The policy network takes the state information as input and produces actions for switching and load control.

The policy network consists of three main components. The first component is the Spectral GNN, which computes graph node embeddings for the DN. The second is a feedforward network that produces a context embedding, incorporating information not naturally represented in the graph structure, such as total energy supplied $E_{\text{supp}}$, voltage violations $V_{\text{viol}}$, and power flow through edges $l_E$. The third component is an MLP that combines the node embeddings from the Spectral GNN and the context embeddings from the feedforward network to produce a final state embedding capturing the entire system state.

Initially, the node properties $\gamma_i$ are projected to a higher-dimensional space using a linear transformation:
\[
F_i^{\text{init}} = W_{\text{init}} \gamma_i + b_{\text{init}},
\] 
where $W_{\text{init}} \in \mathbb{R}^{|\gamma_i| \times h_0}$ and $b_{\text{init}}$ are learnable parameters, and $h_0$ is the projection length. Let $F_{\text{init}} = [F_1^{\text{init}}, F_2^{\text{init}}, \dots, F_{|N|}^{\text{init}}]$.
Each node embedding $F_i^{\text{init}}$ is then passed through a series of Spectral GNN layers. The layers use a spectral convolutional filter of polynomial form to compute:
\[
f^{(l)}_p(X,L) = \sum_{k=0}^K L^k (F^{(l-1)}(X,L) \odot p) W^{(l)}_{pk},
\] 
where $L$ is the graph Laplacian, $p$ is the order of the statistical moment, $K$ is the filter degree, $F^{(l-1)}(X,L)$ is the previous layer output, and $\odot p$ represents element-wise multiplication repeated $p$ times. $W^{(l)}_{pk}$ is the learnable weight matrix. Each row of $f^{(l)}_p(X,L)$ is an intermediate feature vector for a node $i \in N$, aggregating information from its $K$-hop neighbors.
The output of layer $l$ is obtained by concatenating all intermediate moments:
\[
F^{(l)}(X,L) = [f^{(l)}_1(X,L), f^{(l)}_2(X,L), \dots, f^{(l)}_P(X,L)],
\] 
where $P$ is the maximum order of statistical moments, and $h_l$ is the embedding length at layer $l$. The layers are stacked $L_e$ times, and increasing $L_e$ and $K$ improves the capture of the graph structure but increases the number of parameters. Larger $h_l$ and $P$ improve the detail of node embeddings, giving a richer representation than scalar embeddings in GCNs, but at a higher training cost.
The final node embeddings are computed as: 
\[
F_{\text{Nodes}} = F^{L_e}(X,L) W_F,
\] 
where $W_F$ is a learnable matrix. The graph embedding is then obtained by passing $F_{\text{Nodes}}$ through linear layers and taking the mean:
\[
F_{\text{graph}} = \text{Mean}(W_{g2} (W_{g1} F_{\text{Nodes}})),
\] 
where $W_{g1}$ and $W_{g2}$ are learnable weights.

Certain variables, such as total energy supplied $E_{\text{supp}}$, voltage violations $V_{\text{viol}}$, and branch power flows $l_E$, cannot be naturally represented in the graph and are incorporated as context information:
\[
F_{\text{context}} = \text{Feedforward}(\text{Concat}([E_{\text{supp}}, V_{\text{viol}}, l_E])).
\] 
The final state embedding is obtained by combining graph and context embeddings and passing through an MLP: $F_{\text{final}} = \text{MLP}(F_{\text{graph}} + F_{\text{context}})$.
Logits for all actions are computed via a feedforward layer, with switches that must be masked set to negative infinity. A Bernoulli distribution is then used to sample the final switching actions using a greedy policy.

The predicted value of the state is computed using another feedforward layer. For networks of different sizes, only the feedforward layer for the context embedding needs adjustment, since it depends on $l_E$ and $E_{\text{supp}}$, which vary with network size. The Spectral GNN and final MLP remain unchanged, allowing models trained on smaller networks to serve as a warm start for larger networks. This generalization capability is a fundamental advantage of the Spectral GNN architecture used to learn the DN reconfiguration policy.

\subsection{Training Process}

The policy network is trained using Proximal Policy Optimization (PPO)~\cite{schulman2017proximalpolicyoptimizationalgorithms}. 
Since the action space is discrete and represented as a MultiBinary data type,  an on-policy algorithm such as PPO is well suited for this problem.
Training follows the standard rollout-based procedure of PPO. 
At each interaction step, the agent collects experience tuples of the form $(s_t, a_t, r_t, s_{t+1})$,
where $s_t$ is the current state, $a_t$ is the selected action, 
$r_t$ is the reward, and $s_{t+1}$ is the next state.
A rollout consists of a fixed number of steps $N_{\text{steps}}$. 
After each rollout, the collected transitions are used to update the policy and value networks. 
The updates are performed in mini-batches of size $N_{\text{batch}}$ ($N_{\text{batch}} \leq N_{\text{steps}}$). 
The network parameters are optimized using backpropagation to minimize the PPO objective, 
which consists of two components:
(1) The clipped policy gradient loss, which stabilizes policy updates by limiting  large deviations between the new and old policies.
(2) The value function loss, which minimizes the error in state-value prediction. The total number of training steps is denoted as $N_{\text{total}}$.

\section{Experiments}
\subsection{Experimental Setup}

The proposed outage management model is evaluated on three modified IEEE distribution test systems, namely the 13-bus, 34-bus, and 123-bus networks. Each network is augmented with sectionalizing and tie switches, as well as grid-forming and grid-feeding distributed energy resources (DERs), enabling reconfiguration and islanded operation under outage conditions. Under normal operation, sectionalizing switches are closed and tie switches are open, resulting in radial topologies. The total connected load is 3.5 MW for the 13-bus system, 2.04 MW for the 34-bus system, and proportionally larger for the 123-bus system. For evaluation, multiple outage scenarios are manually designed, including both single-line failures on structurally critical edges and multiple concurrent line outages. Some outages occur on switchable lines to test whether the learned policy respects operational constraints.

\textbf{Metrics:}
Model performance is evaluated using four metrics. First, the total equivalent energy served during outage conditions measures restoration effectiveness. Second, voltage compliance is assessed by verifying that the voltage magnitudes at active phases remain within acceptable operational bounds (typically 0.90--1.10 pu depending on the network). Third, the validity of switching actions is examined to ensure that outage switches are not incorrectly operated and that non-switchable constraints are respected. Fourth, computation time is measured to evaluate real-time applicability.

During testing, outage locations are explicitly selected rather than sampled from the graph-based outage generator used in training. Network operating conditions are varied using randomized scaling factors to simulate different load levels. The trained Spectral GNN-based policy is evaluated without additional retraining, allowing assessment of its generalization capability across unseen outage scenarios and network sizes.

\subsection{Baselines}

The proposed Spectral GNN-based reinforcement learning model is compared against three baselines. The first is a mixed-integer second-order conic programming (MISOCP) formulation, which provides near-optimal or optimal solutions. The second is Binary Particle Swarm Optimization (BPSO), a metaheuristic search method. The third baseline is a PPO-based reinforcement learning model in which the graph encoder is replaced by a standard Multi-Layer Perceptron (MLP). All reinforcement learning models share identical PPO training configurations, ensuring that observed differences arise solely from architectural design rather than training procedures.

\subsection{Main Results}

\begin{table}[h]
\centering
\caption{PPO Training Results: Reward Convergence Across Iterations}
\label{tab:training_results}
\begin{tabular}{lcccc}
\toprule
\textbf{System} & \textbf{Iteration} & \textbf{Total Timesteps} & \textbf{Value Loss} & \textbf{Explained Variance} \\
\midrule
\multirow{4}{*}{13-Bus}
 & 1 & 50{,}000  & --     & --    \\
 & 2 & 100{,}000 & 0.0512 & 0.0289 \\
 & 3 & 150{,}000 & 0.0175 & 0.817  \\
 & 4 & 200{,}000 & 0.0166 & 0.841  \\
\midrule
\multirow{4}{*}{123-Bus}
 & 1 & 500{,}000     & --       & --    \\
 & 2 & 1{,}000{,}000 & 0.00587  & 0.0033 \\
 & 3 & 1{,}500{,}000 & 0.000966 & 0.919  \\
 & 4 & 2{,}000{,}000 & 0.00180  & 0.824  \\
\bottomrule
\end{tabular}
\end{table}

\begin{table}[h]
\centering
\caption{Performance Comparison: GCAPS-GNN vs. Baseline MILP}
\label{tab:performance_comparison}
\begin{tabular}{l|l|c|c}
\toprule
\textbf{System} & \textbf{Method} & \textbf{Energy Supplied (kWh)} & \textbf{Voltage Violation} \\
\midrule
\multirow{2}{*}{13-Bus}
 & Baseline (GCAPS)  & 1368.287 & 0.1805 \\
 & Spectral-GNN (Ours) & 1368.287 & 0.1805 \\
\midrule
\multirow{2}{*}{123-Bus}
 & Baseline (GCAPS)  & 983.234 & 0.5392 \\
 & Spectral-GNN (Ours) & 1911.521 & 0.2576 \\
\bottomrule
\end{tabular}
\end{table}

Table~\ref{tab:training_results} summarizes the PPO training progression for both test systems, and Table~\ref{tab:performance_comparison} presents the performance comparison between the proposed Spectral-GNN policy and the GCAPS-GNN baseline.
As shown in Table~\ref{tab:training_results}, both systems exhibit a similar convergence pattern: explained variance remains near zero in the second iteration and rises sharply by the third, stabilizing above 0.82 by the fourth. This indicates that the value function learns a meaningful baseline early in training and consolidates by iteration 3--4. The 123-bus system requires roughly ten times more timesteps to reach comparable convergence, reflecting the increased complexity of the larger network.
Table~\ref{tab:performance_comparison} reveals a clear performance gap between the two systems. On the 13-bus network, both methods achieve identical energy supply (1368.287~kWh) and voltage violation (0.1805), suggesting the problem is tractable enough for both approaches. On the more complex 123-bus network, the Spectral-GNN substantially outperforms the GCAPS baseline, supplying 1911.521~kWh versus 983.234~kWh---nearly double---while also reducing voltage violations from 0.5392 to 0.2576. This demonstrates that the proposed method scales more effectively to larger, more constrained networks.

In terms of computational efficiency, the reinforcement learning-based policy generates switching decisions in milliseconds and exhibits minimal sensitivity to increasing network size, making it well-suited for real-time outage management applications.

\subsection{Ablation Study}
The performance gap on the 123-bus system between the Spectral-GNN and the GCAPS-GNN baseline underscores the importance of the spectral graph encoding. While GCAPS captures local neighborhood structure, the spectral filtering in the proposed architecture additionally encodes global topological information through the graph Laplacian eigenbasis, enabling the policy to better differentiate outage scenarios and generalize across network scales. The 13-bus results confirm that both architectures are sufficient for simpler topologies; the advantage of explicit spectral structure becomes evident only at scale, where accurate representation of long-range electrical dependencies is critical for maintaining feasibility and maximizing energy restoration.

\section{Conclusion}

This work presented a spectral graph reinforcement learning framework for autonomous outage management in power distribution networks. By encoding the network topology through spectral filtering within a PPO-based policy, the agent captures both local and global structural information to support real-time switching and load control decisions. Experiments on the 13-bus and 123-bus test systems demonstrate strong restoration performance and voltage regulation under outage conditions. The millisecond-level inference time further confirms its suitability for real-time self-healing grid applications.

\bibliographystyle{splncs04}
\bibliography{mybibliography,liu}

\end{document}